\documentclass{article}
\usepackage{spconf,amsmath,graphicx}
\usepackage{bm}
\usepackage{globalvals}  
\usepackage{here}
\usepackage{subcaption}

\defVal{tab_col_space}{0.35cm}
\defVal{tab_col_space_s}{0.25cm}
\defVal{fig_1col_width}{0.8}
\defVal{fig_2col_width}{0.95}
\defVal{vspace_under_fig}{-0.2cm}

\title{Weakly Supervised Instance Segmentation \\ using Motion Information via Optical Flow}
\name{Jun Ikeda \qquad Junichiro Mori}
\address{The University of Tokyo, 7--3--1 Hongo, Bunkyo-ku, Tokyo, Japan \\
         \normalsize ji@g.ecc.u-tokyo.ac.jp, mori@mi.u-tokyo.ac.jp}
\begin{document}
%
\maketitle
\begin{abstract}
Weakly supervised instance segmentation has gained popularity because it reduces
high annotation cost of pixel-level masks required for model training.
Recent approaches for weakly supervised instance segmentation detect and segment objects using appearance information obtained from a static image. However, it poses the challenge of identifying objects with a non-discriminatory appearance.
In this study, we address this problem by using motion information from image sequences. 
We propose a two-stream encoder that leverages appearance and motion features extracted from images and optical flows.
Additionally, we propose a novel pairwise loss that considers both appearance and motion information to supervise segmentation.
We conducted extensive evaluations on the YouTube-VIS 2019 benchmark dataset.
Our results demonstrate that the proposed method improves the Average Precision of the state-of-the-art method by 3.1.
\end{abstract}
\section{Introduction}
\label{sec:introduction}
Instance segmentation, which is a task of detecting objects in an image and segmenting the region of the objects at the pixel level, is one of the key technologies behind recently advanced machine vision applications.
It has been actively studied because of its wide range of applications, such as autonomous driving.
However, the annotation cost of pixel-level masks required to train models has been one of the major challenges.
Therefore, weakly supervised instance segmentation, which can train models without relying on mask annotations, has recently gained popularity.

\begin{figure}[t]
    \centering
    \includegraphics[width=\useVal{fig_2col_width}\linewidth]{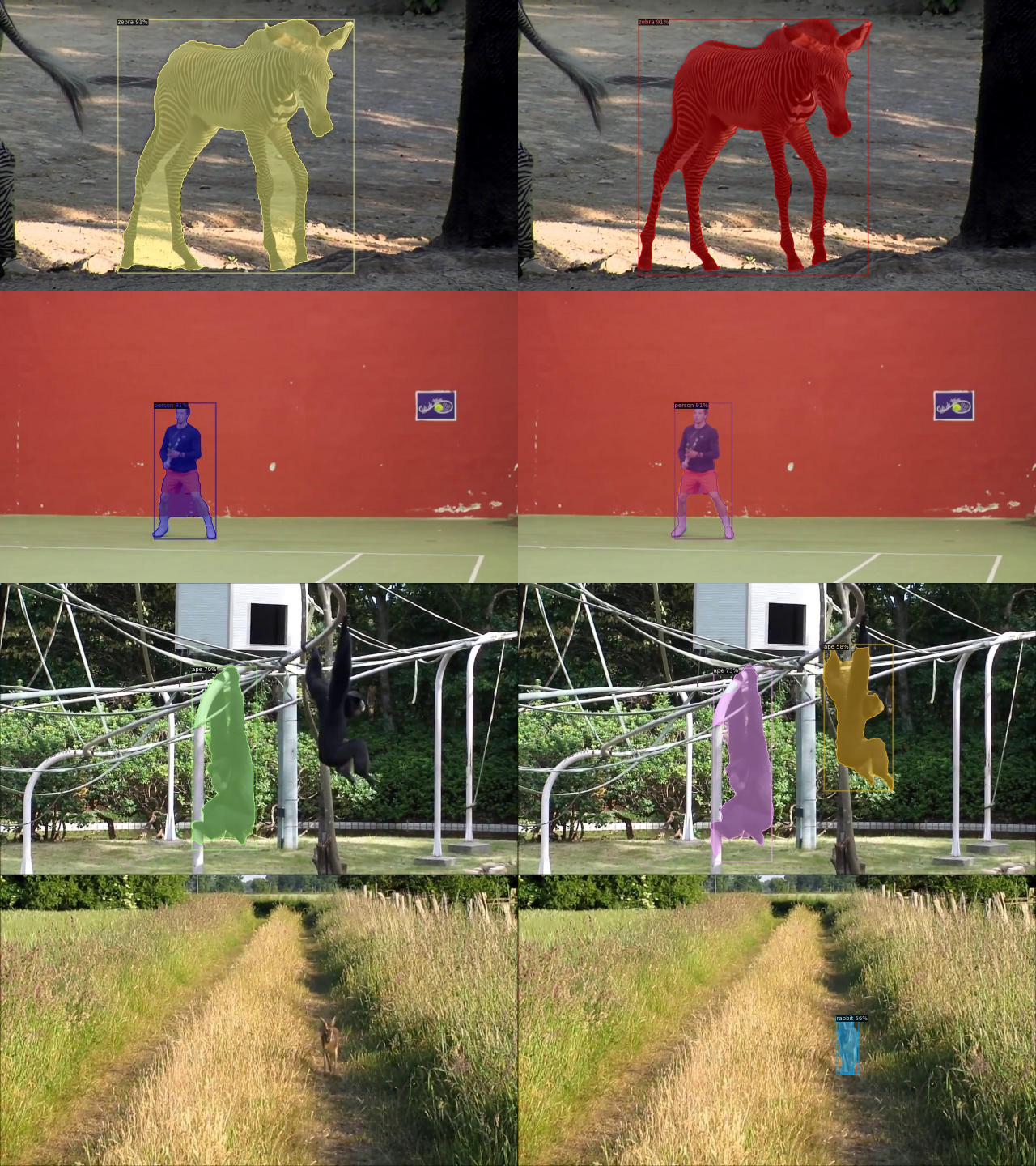}
    \caption{Some example of instance segmentation and object detection results: the state-of-the-art method~\cite{Tian_2021_CVPR_boxinst} (left) and the proposed method (right). Our proposed method reduces the errors observed in the results of the state-of-the-art method.}
    \label{fig:results}
    \vspace{\useVal{vspace_under_fig}}
\end{figure}

Box-supervised instance segmentation is a common weakly supervised approach in which a model is trained using information from bounding boxes as pseudo-labels for the segmentation mask.
The main challenge in weakly supervised mask generation with box annotations is the separation of foreground as an object and its background regions in the box. If the entire box region is labeled as foreground, many false positive labels adversely affect the model training for mask generation.
In previous works~\cite{Khoreva_2017_CVPR_simple_does_it, Hsu_2019_NIPS_BBTP, Lee_2021_CVPR_BBAM, Tian_2021_CVPR_boxinst}, foreground is separated based on image appearance information.
For example, BoxInst~\cite{Tian_2021_CVPR_boxinst}, which is the state-of-the-art approach, has addressed the separation problem by segmenting regions based on the color similarity of local pixel pairs.
However, it suffers from segmentation errors, which caused by false connections of regions with similar colors in the foreground and background (Figure~\ref{fig:results} upper left).
Additionally, it often fails to detect objects with non-discriminative appearances (Figure~\ref{fig:results} lower left).

While previous studies have solely relied on information from a static image, they ignored information from image sequences. In fact, most segmentation technology applications, such as autonomous driving and video editing, use image sequences as system inputs.

In this study, we address the above-mentioned problems by introducing temporal information from sequential images as complement signals for weakly supervised instance segmentation.
In particular, we propose a method for weakly supervised instance segmentation using motion information.
The basic idea is to use motion information in a form of optical flow~\cite{Baker_2007_ICCV_optical_flow}, which could support object detection and segmentation when dealing with a foreground object moving differently than background, as shown in Figure~\ref{fig:results}.

It has been demonstrated that motion information is effective in achieving the task of so-called motion segmentation~\cite{Jain_2017_CVPR_FusionSeg, Siam_2018_ITSC_MODNet, Rashed_2019_ICCV_FuseMODNet, Mohamed_2021_IV_instance_motion_segmentation}, which attempts to segment the region of moving objects in a class-agnostic manner.
In the motion segmentation, motion feature is extracted from optical flow using a pre-trained object detection model~\cite{Jain_2017_CVPR_FusionSeg} as its backbone. 
Previous studies showed that integrating motion feature maps into appearance feature is effective~\cite{Siam_2018_ITSC_MODNet}.
Moreover, further fine-grained architectures, such as multi-level feature maps~\cite{Rashed_2019_ICCV_FuseMODNet} and feature pyramid networks~\cite{Mohamed_2021_IV_instance_motion_segmentation}, have recently been introduced. 

Several recent approaches for the instance segmentation have also explored the use of temporal information by aggregating features across frames~\cite{Fu_2021_AAAI_CompFeat, Bertasius_2020_CVPR_MaskProp} or using 3D convolutions~\cite{Voigtlaender_2019_CVPR_mots, Athar_2020_ECCV_StemSeg}. 
Previous work~\cite{Liu_2021_CVPR_temporal_mask_consistency} attempted to use motion information for instance segmentation in a weakly supervised manner. However, it simply amplifies the foreground scores of the regions with large motion. 

We propose a two-stream encoder that uses motion features extracted from optical flow for object detection and segmentation, with the goal of weakly supervised instance segmentation using motion information.
Additionally, we propose a novel pairwise loss that uses optical flow to complement the segmentation of foreground and background regions with similar colors.
The contributions of this study are summarized as follows:(1) 
We explore the use of motion information in box-supervised instance segmentation. (2) We propose a two-stream encoder to provide both appearance and motion features for the detection and mask heads. Additionally, we propose a novel pairwise loss that uses both appearance and motion information to improve segmentation accuracy. (3) Extensive evaluation on the YouTube-VIS 2019~\cite{yang_ICCV_2019_VIS} benchmark dataset demonstrates that the proposed method outperforms the state-of-the-art approach even with less complex pre-trained backbones.

\section{Motion-aided Instance Segmentation}
Our method for weakly supervised instance segmentation leverages motion information on top of the architecture of BoxInst~\cite{Tian_2021_CVPR_boxinst} which is a box-supervised instance segmentation approach.
BoxInst employs projection and pairwise losses to supervise mask learning in CondInst~\cite{Tian_2020_ECCV_CondInst} with box annotations.
CondInst is a high-performance instance segmentation model comprised FCOS~\cite{Tian_2019_ICCV_FCOS}, an anchor-free object detection method, and an instance-aware mask head whose parameters are generated dynamically for each object.
In BoxInst, the projection loss is used to train the mask head. It supervises the horizontal and vertical projections of the predicted mask using the ground-truth box annotation.
Furthermore, the pairwise loss encourages neighboring pixel pairs with similar colors to have the same label, allowing the pseudo-mask labels to propagate.

\begin{figure}
    \centering
    \includegraphics[width=\useVal{fig_2col_width}\linewidth]{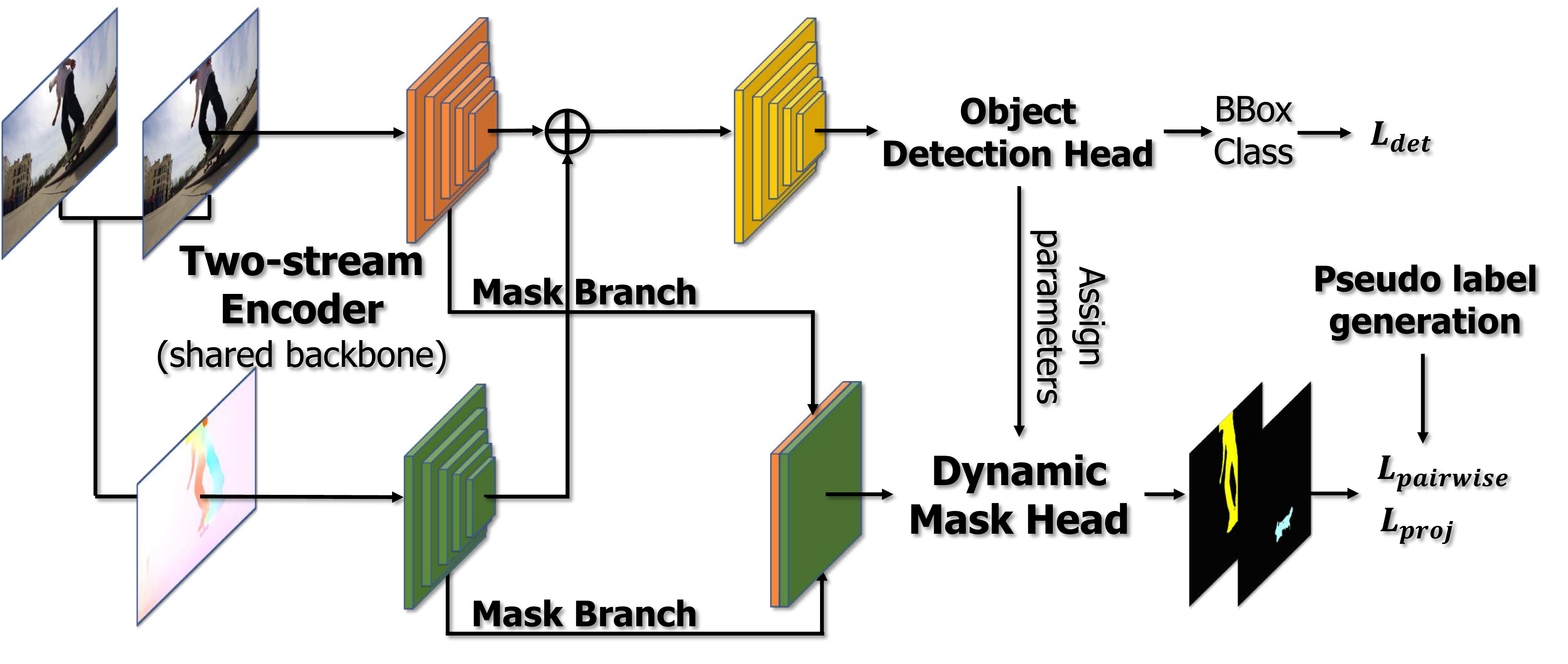}
    \caption{Overview of our model, which includes a two-stream encoder for extracting appearance and motion feature maps, which are then fed into the detection and instance-aware mask heads. The projection and the proposed novel pairwise losses are used to supervise the mask head.}
    \label{fig:architecture}
    \vspace{\useVal{vspace_under_fig}}
\end{figure}

To use motion information as complement signals for the weak supervision, we use a two-stream encoder to provide suitable feature maps for the detection and mask heads, respectively, on top of the box-supervised architecture. Additionally, we introduce a novel pairwise loss to supervise the mask head. 
Figure~\ref{fig:architecture} shows the overall architecture of our proposed model.
The architecture consists of the two-stream encoder, object detection head, and dynamic mask head.

\subsection{Two-stream encoder}
\label{ssec:two-stream_encoder}
Our two-stream encoder first extracts appearance and motion features from the input image and its optical flow, respectively.
We assume that the input image $\bm{I}_{t}$ has box annotations, and its optical flow, in RGB color space, is estimated from the image sequence $\bm{I}_{t}$ and $\bm{I}_{t+1}$, where $t$ denotes the time-step of a target frame. Each stream consists of a backbone network pre-trained on ImageNet~\cite{Deng_2009_CVPR_ImageNet} and FPN~\cite{Lin_2017_CVPR_FPN} to extract multi-level feature maps $\bm{F}_\mathrm{img}^P$ and $\bm{F}_\mathrm{flow}^P$ ($P \in (3,4,5,6,7)$ represents the level of the feature maps).
The parameters of the backbone network are shared between the two streams.

The appearance and motion features are then combined to form a feature map, which is then fed into the detection and mask heads, respectively.
 $\bm{F}_\mathrm{img}^P$ and $\bm{F}_\mathrm{flow}^P$ are fused to generate the feature map $\bm{F}_\mathrm{det}^{P}$ to be fed into the detection head using the following summation function: $\bm{F}_\mathrm{det}^{P} = \bm{F}_\mathrm {img}^{P} + \bm{F}_\mathrm{flow}^{P}$. 
This is because appearance and motion features can provide complementary information in object detection.
Appearance features are essential for object classification, whereas motion features can help detect objects with non-discriminative appearance.

For the feature map $\bm{F}_\mathrm{mask}$ to be fed into the mask head, we first obtain $\bm{F}_\mathrm{mask}^\mathrm{img}$ and $\bm{F}_\mathrm{mask}^\mathrm{flow}$ by feeding $\bm{F}_\mathrm{img}^{3}$ and $\bm{F}_\mathrm{flow}^{3}$ into the mask branch~\cite{Tian_2020_ECCV_CondInst}, and then into the mask head.
The mask branch consists of four $3 \times 3$ convolutions with 128 channels before the last layer and $C_\mathrm{mask}$ channels for the last layer.
As a result, the number of parameters in the instance-aware mask head is reduced.
Following that, $\bm{F}_\mathrm{mask}^\mathrm{img}$ and $\bm{F}_\mathrm{mask}^\mathrm{flow}$ are concatenated in the channel direction to obtain $\bm{F}_\mathrm{mask}$.
Concatenating feature maps rather than adding them is done to take advantage of the instance-aware mask head, which enables the model to use appearance and motion features selectively for segmentation depending on the object.

\subsection{Pairwise loss}
\label{ssec:pairwise_loss}
Our newly introduced pairwise loss is designed to take motion information into account to more accurately impose the pseudo-mask label-identity constraint, with the color similarity of local pixel pairs.
To begin, from the score map $\bm{m}$ output by the mask head, the probability that a pixel pair $e$ has the same label can be determined using the following equation:
\begin{equation}\label{eq:label_identity_prob}
    P(y_e = 1) = m_{i,j} \cdot m_{k,l} + (1 - m_{i,j}) \cdot (1 - m_{k,l}),
\end{equation}
where $y_e$ is the pairwise identity label, being one if the pixels belonging to $e$ have the same label and 0 otherwise.
Considering $m$ as the probability that the corresponding pixel has the foreground label, Equation~\ref{eq:label_identity_prob} represents the probability that the two pixels, $(i,j)$ and $(k,l)$, have a common label (e.g., the foreground).
Then, the pairwise loss is expressed as follows:
\begin{equation}\label{eq:pairwise_loss}
    L_\mathrm{pairwise} = - \frac{1}{N} \sum_{e \in E_\mathrm{in}} y_e \log P(y_e = 1),
\end{equation}
where $E_\mathrm{in}$ the set of pixel pairs with at least one pixel is in the box. $N$ is the number of pairs in $E_\mathrm{in}$.

Because the ground truth label-identity $y_e$ is not available, we estimate the label-identity using the appearance and motion information obtained from the input image and optical flow.
The pseudo label-identity is defined by comparing the color similarity $S_e^\mathrm{color}$ and optical flow similarity $S_e^\mathrm{flow}$ of each pixel pair $e$.
The similarities,  $S_e^\mathrm{color}$ and $S_e^\mathrm{flow}$, are defined as follows:
\begin{equation}
    S_e^\mathrm{color} = S^\mathrm{color} (\bm{c}_{i,j}, \bm{c}_{k,l}) = \exp \left(-\frac{||\bm{c}_{i,j}-\bm{c}_{k,l}||}{\theta_\mathrm{color}}\right),
\end{equation}
\begin{equation}
    S_e^\mathrm{flow} = S^\mathrm{flow} (\bm{f}_{i,j}, \bm{f}_{k,l}) = \exp \left(-\frac{||\bm{f}_{i,j}-\bm{f}_{k,l}||}{\theta_\mathrm{flow}}\right),
\end{equation}
where $\bm{c}$ is a color vector in Lab color space and $\bm{f}$ is an optical flow vector in RGB color space.
$\theta_\mathrm{color}$ and $\theta_\mathrm{flow}$ are hyper-parameters.
Finally, the pseudo label-identity is written as $y_e = 1$ if $S_e^\mathrm{color} \geq \tau_\mathrm{color}$ and $S_e^\mathrm{flow} \geq \tau_\mathrm{flow}$, and $y_e = 0$ otherwise, where $\tau_\mathrm{color}$ and $\tau_\mathrm{flow}$ are pre-defined thresholds. 

\section{Experiments and Results}
\label{sec:experiments}

\subsection{Dataset and Implementation Details}
We conducted our experiments using the YouTube-VIS 2019~\cite{yang_ICCV_2019_VIS} benchmark dataset.
Because the annotations of the original val split are not publicly available, we randomly selected ten videos per class from the original train split, following~\cite{Liu_2021_CVPR_temporal_mask_consistency}, to create \textit{train\_val} split, and named the remaining data \textit{train\_train} split.
Finally, the \textit{train\_train} split contains 1847 videos and 51049 images, while the \textit{train\_val} split contains 391 videos and 10796 images.
The images in the \textit{train\_train} split are used to train our model. The model is evaluated on the images and mask annotations in the \textit{train\_val} split. The evaluation metrics are calculated following the standard procedures defined in the COCO dataset~\cite{Lin_ECCV_2014_coco}.
For optical flow estimation, we use the recent pre-trained model~\cite{Jiang_2021_ICCV_GMA} (GMA) on the sintel dataset~\cite{Butler_2012_ECCV_Sintel}, which is  computationally  much less expensive.
Finally, we train our model for 90K iterations with a batch size of 16 on eight A100 GPUs.
In our setting, $\tau_\mathrm{flow} = 0.6$ and $\theta_\mathrm{flow} = 0.5$. The other hyperparameters are the same as in BoxInst~\cite{Tian_2021_CVPR_boxinst}.

\subsection{Instance Segmentation Results}
We compare the instance segmentation results of the proposed method with BoxInst~\cite{Tian_2021_CVPR_boxinst} as the baseline.
The proposed method, which used ResNet-50 as its backbone, improves the baseline by 3.1 in Average Precision (AP) and by 4.0 in $\mathrm{AP_{75}}$ (Table~\ref{tab:results_IS}).
In fact, our qualitative comparison of the instance segmentation results, as shown in the upper half of Figure~\ref{fig:results}, demonstrate that the proposed method greatly reduces the false positive errors observed in the results of the baseline.
These results indicate that incorporating the motion information improves the accuracy of segmentation of ambiguous color regions, which makes it difficult to separate the foreground from the background.

\begin{table}[t]
   \caption{The results of instance segmentation}
    \label{tab:results_IS}
    \centering
    \begin{tabular}{lllll}
        \hline
        Method & Backbone & AP & $\mathrm{AP}_{50}$ & $\mathrm{AP}_{75}$ \\ \hline
        BoxInst & ResNet-50 & 25.9 & 46.8 & 25.4 \\
        Ours & ResNet-50 & \textbf{29.0} & \textbf{50.2} & \textbf{29.4} \\ \hline
        BoxInst & ResNet-101 & 29.1 & 51.2 & 29.4 \\
        Ours & ResNet-101 & \textbf{30.1} & \textbf{52.0} & \textbf{30.6} \\ \hline
    \end{tabular}
    \vspace{\useVal{vspace_under_fig}}
\end{table}

\begin{table}[t]
    \caption{The results of ablation study on model components}
    \label{tab:ab_study}
    \centering
    \begin{tabular}{l@{\hspace{\useVal{tab_col_space}}}l@{\hspace{\useVal{tab_col_space}}}l@{\hspace{\useVal{tab_col_space}}}l}
        \hline
        Method & AP & $\mathrm{AP}_{50}$ & $\mathrm{AP}_{75}$ \\ \hline
        w/o motion feat. for detection & 27.9 & 48.2 & 29.0 \\
        w/o motion feat. for segmentation & 27.6 & 48.8 & 27.7 \\
        w/o optical flow for pairwise loss & 27.3 & 48.8 & 27.3 \\ \hline
        Ours & 29.0 & 50.2 & 29.4 \\ \hline
    \end{tabular}
\end{table}

Moreover, we conduct an ablation study to evaluate the contribution of each component of the proposed model. 1) motion feature for detection, 2) motion feature for segmentation, and 3) optical flow for pairwise loss.
The relative decreases in AP and $\mathrm{AP_{75}}$ indicate that introducing motion information to both features for segmentation and pseudo-labels for pairwise loss contributes to the improvement (Table~\ref{tab:ab_study}).
Additionally, it turns out that using motion features for detection is necessary to improve the results.
This is because improving detection quality eventually contributes to better instance segmentation results.

\subsection{Objection Detection Results}
Table~\ref{tab:results_OD} shows the comparative results for object detection performance.
The proposed method with ResNet-50 as its backbone improves the baseline by 1.4 in AP.
As our successful cases shown in the lower half of Figure~\ref{fig:results} demonstrate, the proposed method considerably reduces even the false negative errors observed in the results of the baseline.
\begin{table}[t]
    \caption{The results of object detection}
    \label{tab:results_OD}
    \centering
    \begin{tabular}{lllll}
        \hline
        Method & Backbone & AP & ${\rm AP}_{50}$ & ${\rm AP}_{75}$ \\ \hline
        BoxInst & ResNet-50 & 37.0 & 54.2 & 38.7 \\
        Ours & ResNet-50 & \textbf{38.4} & \textbf{55.6} & \textbf{40.8} \\ \hline
        BoxInst & ResNet-101 & \textbf{39.9} & 56.5 & \textbf{42.3} \\
        Ours & ResNet-101 & 39.7 & \textbf{57.2} & 41.6 \\ \hline
    \end{tabular}
    \vspace{\useVal{vspace_under_fig}}
\end{table}

\section{Discussions}
\subsection{Feature Fusion Comparison}
In our proposed two-stream encoder, the appearance and motion feature maps for the detection head are added together, whereas for the mask head, they are concatenated.
We conduct an additional ablation study to verify each feature fusion strategy by varying the fusion method.
The fusion methods can be as simple as summing and concatenating or as complex as attention.
However, in this study, we compare the methods that do not have a significant effect on memory efficiency or computational cost.
First, we compare the detection head fusion methods, namely, maximum and summation fusion.
As shown in Table~\ref{tab:fusion_functions}, summation-based fusion is superior in the detection results except $\mathrm{AP_L}$.
This suggests the necessity of considering both appearance and motion features to detect objects. Next, we compare the fusion methods for the mask heads, that is maximum, summation, and concatenation fusion.
As a result, the concatenation approach is the best for AP, $\mathrm{AP_{50}}$, $\mathrm{AP_M}$ and $\mathrm{AP_L}$.
However, for $\mathrm{AP_{75}}$ and $\mathrm{AP_S}$, the maximum approach is superior.
Therefore, for small objects, it is preferable to use the maximum signals of both features.
\begin{table}[t]
\caption{The results of ablation study on fusion methods}
\label{tab:fusion_functions}
\begin{minipage}[b]{1.0\linewidth}
    \subcaption{Fusion methods for the detection head.}\label{tab:sub_fusion_for_detection}
    \centering
    \begin{tabular}{l@{\hspace{\useVal{tab_col_space_s}}}l@{\hspace{\useVal{tab_col_space_s}}}l@{\hspace{\useVal{tab_col_space_s}}}l@{\hspace{\useVal{tab_col_space_s}}}l@{\hspace{\useVal{tab_col_space_s}}}l@{\hspace{\useVal{tab_col_space_s}}}l}
        \hline
        Fusion method & AP & ${\rm AP}_{50}$ & ${\rm AP}_{75}$ & ${\rm AP}_{\rm S}$ & ${\rm AP}_{\rm M}$ & ${\rm AP}_{\rm L}$ \\ \hline
        maximum & 37.7 & 54.6 & 40.4 & 8.5 & 17.5 & \textbf{40.7} \\
        summation & \textbf{38.4} & \textbf{55.6} & \textbf{40.8} & \textbf{9.2} & \textbf{18.2} & 40.3 \\ \hline
    \end{tabular}
\end{minipage}
\vspace{0.2cm} \\
\begin{minipage}[b]{1.0\linewidth}
    \subcaption{Fusion methods for the mask head.}\label{tab:sub_fusion_for_mask}
    \centering
    \begin{tabular}{l@{\hspace{\useVal{tab_col_space_s}}}l@{\hspace{\useVal{tab_col_space_s}}}l@{\hspace{\useVal{tab_col_space_s}}}l@{\hspace{\useVal{tab_col_space_s}}}l@{\hspace{\useVal{tab_col_space_s}}}l@{\hspace{\useVal{tab_col_space_s}}}l}
        \hline
        Fusion method & AP & $\mathrm{AP_{50}}$ & $\mathrm{AP_{75}}$ & $\mathrm{AP_S}$ & $\mathrm{AP_M}$ & $\mathrm{AP_L}$ \\ \hline
        maximum & 28.6 & 49.3 & \textbf{29.6} & \textbf{5.1} & 9.7 & 31.0 \\
        summation & 28.3 & 49.0 & 29.2 & 3.2 & 10.0 & 31.1 \\
        concat & \textbf{29.0} & \textbf{50.2} & 29.4 & 3.8 & \textbf{11.8} & \textbf{31.6} \\ \hline
    \end{tabular}
\end{minipage}
\end{table}

\subsection{Visualizing Activation Map }
\begin{figure}[t]
    \centering
    \includegraphics[width=\useVal{fig_2col_width}\linewidth]{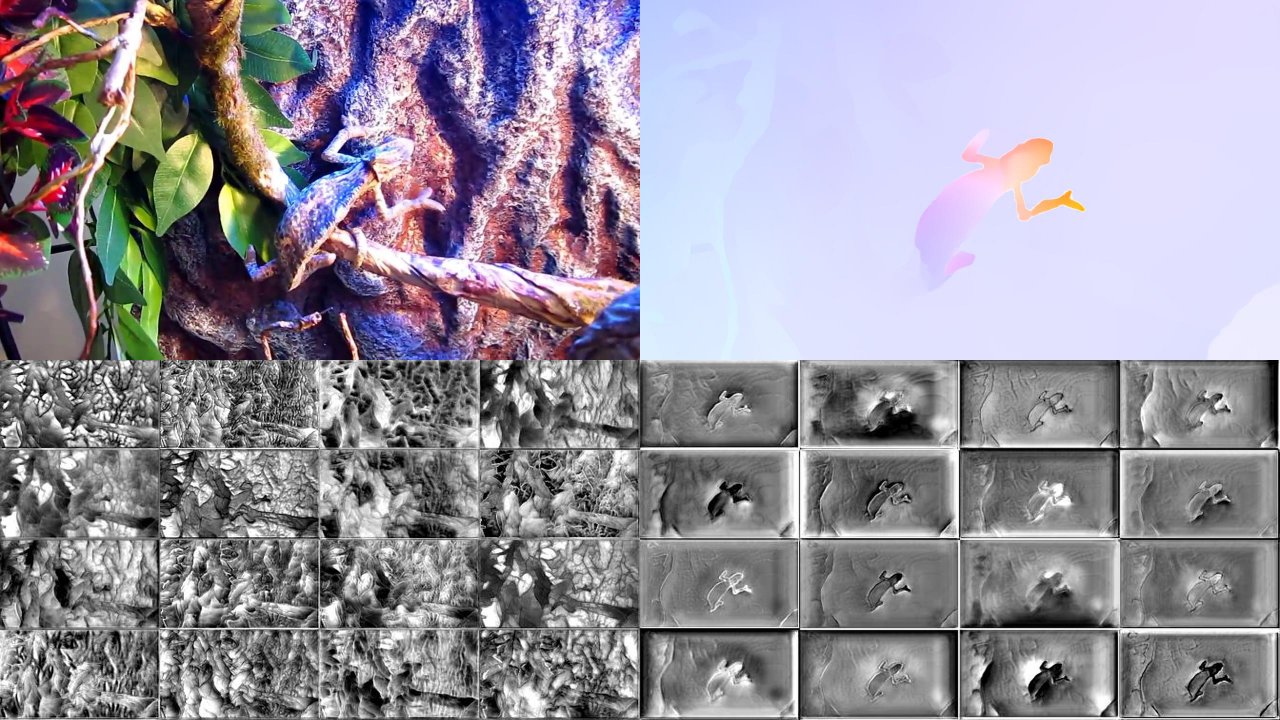}
    \caption{The activation maps, which are fed into the mask head, of the appearance features (lower left) and motion features (lower right) corresponding to the input image (upper left) and optical flow (upper right). }
    \label{fig:activation_maps}
    \vspace{\useVal{vspace_under_fig}}
\end{figure}

Figure~\ref{fig:activation_maps} shows the activation maps of the appearance features and motion features . It is worth noting that the motion features can provide discriminatory signals even when the foreground (i.e., a reptile) is blended with its background, where the appearance features do not provide relevant signals. This suggests the importance of motion features as complement signals for box-supervised instance segmentation.

\section{Conclusion and Future Work}
\label{sec:conclusion}
In this study, we propose a two-stream encoder and novel pairwise loss that successfully incorporates both appearance and motion information for weakly-supervised instance segmentation.
Our comparison study demonstrated that the proposed method outperforms the state-of-the-art method in terms of more accurate detection and segmentation of objects with non-discriminative appearance.
Our future work is to consider negative samples of label-identity classification in the pairwise loss.
Learning negative samples, such as label-contrariety, will enable explicit boundary separation.

\subsubsection*{Acknowledgement}
This work was supported by JST, CREST Grant Number JPMJCR21D1 and MEXT/JSPS KAKENHI Grant Number JP20K12076.


\vfill\pagebreak

\bibliographystyle{IEEEbib}
{\small \bibliography{main}}

\end{document}